\documentclass[11pt]{article}
\usepackage{times}
\usepackage{graphicx}
\usepackage{amsmath}
\usepackage{hyperref}
\usepackage{caption}
\usepackage{subcaption}
\usepackage{booktabs}
\usepackage{url}
\usepackage[margin=1in]{geometry}
\usepackage{float}

\title{Explainable Artificial Intelligence Credit Risk Assessment using Machine Learning}

\author{
  Shreya\\
  Department of Computer Science and Engineering \\
  Specialisation in Data Science and Artificial Intelligence \\
  SRM University, Sonepat, Haryana, India \\
  \texttt{shreya.work224@gmail.com}
  \and
  Mr. Harsh Pathak\\
  Assistant Professor \\
  Department of Computer Science and Engineering \\
  SRM University, Sonepat, Haryana, India \\
  \texttt{harshpathak@srmuniversity.ac.in}
}

\date{}

\begin{document}

\maketitle

\begin{abstract}
This paper presents an intelligent and transparent AI-driven system for Credit Risk Assessment using three state-of-the-art ensemble machine learning models combined with Explainable AI (XAI) techniques. The system leverages XGBoost, LightGBM, and Random Forest algorithms for predictive analysis of loan default risks, addressing the challenges of model interpretability using SHAP and LIME. Preprocessing steps include custom imputation, one-hot encoding, and standardization. Class imbalance is managed using SMOTE, and hyperparameter tuning is performed with GridSearchCV. The model is evaluated on multiple performance metrics including ROC-AUC, precision, recall, and F1-score. LightGBM emerges as the most business-optimal model with the highest accuracy and best trade-off between approval and default rates. Furthermore, the system generates applicant-specific XAI visual reports and business impact summaries to ensure transparent decision-making.
\end{abstract}

\section{Introduction}
Credit risk assessment is a very important process in the financial sector, which entails the assessment of the likelihood of a borrower defaulting on their debt.  Conventionally, this is a human process that involves assessment of several financial parameters, such as debt-to-income ratio, credit history, income stability, and other loan-specific parameters. These assessments are typically time-consuming, prone to human error, and non-scalable.

The purpose of this project is to create a strong and transparent AI-based credit risk assessment system to properly predict loan default probabilities. For this, the project utilizes the strengths of three of the most advanced machine learning algorithms: XGBoost, LightGBM, and Random Forest. These algorithms are the first choice for credit risk assessment due to their high predictive power, ability to learn intricate non-linear relationships, and ability to handle varied financial data. Recognizing the greatest importance of transparency in financial decisions, this project introduces XAI (Explainable Artificial Intelligence)” by incorporating SHAP (SHapley Additive exPlanations) and “LIME (Local Interpretable Model-agnostic Explanations)” into the system. These methods provide explanations about each feature’s contribution towards the model's predictions, adding to the transparency and trustworthiness of the system. 

In addition, the project seeks to produce full XAI reports and human-readable reports for each test applicant. These reports will facilitate manual inspection by stakeholders, including financial lending institutions and the clients/applicants themselves, to better comprehend the credit risk assessment process.

\section{Related Work}
In \cite{dura2024impact}, Tudor explored the transformative “impact of Artificial Intelligence (AI) on Credit Risk Assessment and Business Model Transformation in the Financial Sector”, highlighting the leverage of diverse data and blockchain for accurate and secure evaluations.

In \cite{chang2024credit}, Chang et al. demonstrated how XGBoost outperformed other machine learning and deep learning models with an accuracy of 99.4\% in predicting credit card default status using a Credit Card Customer Data. The key predictors identified were “age, income, employment duration, and number of family members”.

In \cite{liu2024kacdp}, Liu and Zhao introduced KACDP, w/c is a “highly interpretable credit default prediction model using Kolmogorov-Arnold Networks (KANs), which outperformed mainstream models in ROC-AUC and F1 Value while providing high interpretability.”

In \cite{hussain2024enhancing}, Nallakaruppan et al. found that AI-driven techniques, including ensemble methods, decision trees, and neural networks, perform better in predictive accuracy and resilience compared to traditional models using data from a large financial organization.

In \cite{addy2024predictive}, Brown's comprehensive review highlighted those predictive analytics enhances the accuracy and efficiency of risk assessments in credit risk management for banks, with machine learning models outperforming traditional ones.

In \cite{nallakaruppan2024credit}, Nallakaruppan et al. showed that Random Forest achieved high accuracy (0.93 and 0.94) on “Peer-to-Peer Lending Platform Data”, with Shapley values and LIME enhancing interpretability.

In \cite{alafeef2024credit}, Mesteri's study on Commercial Banking Credit Registry Data revealed that Logistic Regression without transformation outperformed all other models, and Weights of Evidence transformation did not improve performance.

In \cite{widagdo2023ai}, Kurniawan and Sri Yusriani's literature review on Artificial Intelligence in Credit Risk indicated a consensus that AI/ML provides better forecast power and can increase credit access, with XGBoost emerging as popular after 2020.

In \cite{blom2023value}, Blom et al. estimated “Value at Risk in the EURUSD Currency Cross from Implied Volatilities Using Machine Learning Methods and Quantile Regression”[9] on over-the-counter (OTC) and foreign exchange (FX) market data.

In \cite{hjelkrem2022open}, Hjelkrem et al.'s case study on a Norwegian bank demonstrated that LightGBM significantly outperformed the bank's current Logistic Regression model in application credit scoring using unsecured consumer loans data.

In \cite{yusuff2021credit}, this study Using Predictive Analytics found that Gradient Boosting outperformed “Logistic Regression”, “Decision Trees”, and “Random Forests” on a public dataset.

In \cite{addo2018credit}, Nallakaruppan et al.'s research utilizes a European bank dataset, found that the Gradient Boosting Machine achieved the highest AUC score of 0.994206.

In \cite{islam2009ai}, Islam et al.'s study on German Credit Card Data showed that an Artificial Neural Network (ANN) achieved slightly better accuracy (83.86\%) compared to traditional methods.

\begin{table}[h]
\centering
\caption{Literature Review Summary}
\begin{tabular}{|p{1.2cm}|p{3.5cm}|p{3.5cm}|p{6cm}|}
\hline
\textbf{Title} & \textbf{Dataset} & \textbf{Algorithm} & \textbf{Result} \\
\hline
\cite{dura2024impact} & Not specified & Machine Learning & Uses behavioural and utility data to improve credit risk predictions and trust via blockchain. \\
\hline
\cite{chang2024credit} & Credit Card Customer Data & NN, LR, AdaBoost, XGBoost, LightGBM & XGBoost achieved 99.4\% accuracy. Key features: age, income, employment, dependents. \\
\hline
\cite{liu2024kacdp} & Not specified & KANs, LR, XGBoost, SVM & KANs delivered best results with interpretability using feature attribution. \\
\hline
\cite{hussain2024enhancing} & Financial Institution Data & Ensemble, DT, NN, LR, LDA & AI methods surpassed traditional ones in accuracy and robustness. \\
\hline
\cite{addy2024predictive} & Review of Bank Data & ML, NN, Ensemble, Probit, Logit & ML models more accurate. Zmijewski \& Ohlson excelled in SME. \\
\hline
\cite{nallakaruppan2024credit} & P2P Lending Data & DT, RF, LR, LDA, SHAP, LIME & RF gave highest accuracy (0.93–0.94). XAI enhanced transparency. \\
\hline
\cite{alafeef2024credit} & Bank Credit Registry & LR, RF, DT, SVM, GBM & Raw LR outperformed WoE-transformed versions. \\
\hline
\cite{widagdo2023ai} & 9 Studies & SVM, ANN, XGBoost, Hybrid Trees & AI/ML methods offered better access and accuracy. No clear best model. \\
\hline
\cite{blom2023value} & EUR/USD OTC Data (2009–2020) & LightGBM, CatBoost, NN & LightGBM \& CatBoost delivered best VaR estimates. \\
\hline
\cite{hjelkrem2022open} & Norwegian Bank Loans & LightGBM, LR & LightGBM improved ROC AUC by 17\% over existing model. \\
\hline
\cite{yusuff2021credit} & Financial, Demographic Data & LR, DT, GBM, SHAP & GBM outperformed others with best accuracy and F1-score. \\
\hline
\cite{addo2018credit} & 117K Co. Financial Data & ElasticNet, RF, GBM, DL & GBM had top AUC (0.994), lowest RMSE (0.041). RF closely followed. \\
\hline
\cite{islam2009ai} & German Credit Card Data & ANN, DA, LR & ANN achieved 83.86\% accuracy, better than DA and LR (76.4\%). \\
\hline
\end{tabular}
\label{tab:literature}
\end{table}

\section{Proposed Methodology}
There are separate datasets for training and testing the model. Different csv files are merged with application csv file using the ID as primary key. Data is cleaned, missing values are replaced using Custom Imputation and outliers are handled. Data is transformed using One-Hot Encoding and Standard Scaler in the preprocessing step. After data is formatted to suit the model training phase, Feature Engineering creates new features related to bank needs, financial metrics are calculated required by loan officers to assess the application. This feature engineering allows calculation about applicant’s credit and evaluates its relation with other features in model training. This full dataset is fed to train the machine learning algorithm and learn hidden patterns and relations and classify each applicant as a defaulter (1) or a non-defaulter (0).

After the model is trained to classify in these two classes. The data is further processed in risk assessment module that calculates further financial and business metrics to classify each applicant into Low, Moderate or High-Risk Categories and maps them to a decision whether to approve or reject the applicant. With approval the system calculates approval conditions for the applicant and how it affects the business.

To further enhance the transparency of decision making for both stakeholders and applicants this decision - making factors are processed in the XAI module which helps understand the factors contributing behind the decision making using SHAP and LIME explainability. Finally, a full comprehensive report for each applicant is generated using HTML for viewing to assess situation of each applicant.

\textbf{A.	Data Preprocessing}

The system initiates with a comprehensive preprocessing pipeline to clean, merge and transform financial data of applicants from multiple sources. The data is merged from different sources that highlights applicant details, credit card details, previous loan payment history, external sources reports, bureau reports etc. The pipeline executed is:

\textbf{1.	Custom Imputation: }An imputation class was customed to ensure completeness of the dataset. The imputation is employed to deal with missing values across the numerical and categorical features using different methods. Numerical values are imputed using the median to reduces skew from outliers, while the categorical features are imputed using mode (most frequent entry) as to ensure semantic coherence.

\textbf{2.	Dataset Merging:} Data was acquired from various sources – application data, bureau reports, external reports, previous payment history, credit card history, precious applications etc. All this data is merged using the unique ID assigned to each applicant. For merging, only essential features are exclusively selected. Statistical methods are applied like mean, max, sum, min, count and standard deviation for aggregating features.

\textbf{3.	Outlier Handling:} Outlier handling is done to reduce the risk of overfitting caused by skewed distributions or rare anomalies. Since each variable needs to be assessed independent of the other, clipping method provides a way to mitigate influence of extreme values without discarding the data. Values are capped at the lower and upper bounds beyond three standard deviations from mean for all numerical features. This approach preserves data size and ensure stability for model training.

\textbf{4.	Feature Transformation Pipeline:} Following the custom imputation for numerical and categorical values, a pipeline is constructed to normalise numerical values using StandardScaler and encode categorical values using OneHotEncoder for efficient training using the scikit-learn’s pipeline and ColumnTransformer classes. Constructing a pipeline ensures that preprocessing steps are reproducible for model training.

\textbf{B.	Feature Engineering}
Feature engineering is an essential part to assess credit risk. It creates advanced-domain specific features from raw customer and credit data. Demographic features, financial ratios, aggregated metrics and behavioural flags are engineered to enhance the model interpretability and focus on features essential for risk assessment. Each sub-module handles a specific dataset.

\textbf{C.	Model Selection and Training
}
Three state-of-the-art algorithms are trained with hyperparameter tuning for classifying applicants into defaulters (1) and non-defaulters (0).

\textbf{1.	Random Forest:} Random Forest is a supervised ML algorithm in which multiple decision trees are built collectively (known as ensemble learning) and each decision tree (DT) provides its own prediction. After aggregating outputs of all learners, final prediction is generated. Final prediction is obtained by averaging all predicted outputs when RF is used as a regression problem. This technique of training multiple trees parallelly is referred to as bagging. Each tree in RF is trained on random subset and random features selected at each split of the decision tree, hence referring it a Random Forest. The number of features selected for each DT is limited (not all features are used) so as to make each tree unique. This is to ensure variability and prevent overfitting. Thus, giving more accurate and reliable predictions.

\textbf{2.	XGBoost :} XGBoost is an advanced ensemble learning method that builds decision tress sequentially. It starts by building and training weak learners and improves subsequent trees using the gradients of loss function and hessian derivatives (second-order derivatives) to minimize error with each sequential iteration and build into a strong learner. The algorithm uses regularisation techniques to avoid overfitting. It captures non-linear relationships better than decision trees making it a preferred choice for risk assessment. 

\textbf{3.	LightGBM :} LightGBM is another advanced ensemble learning technique closely similar to XGBoost but with improved speed and scalability. It utilises leaf-wise growth strategy to expand the node with highest possibility of loss reduction  and histogram-binning algorithm to identify  the best split and parallel feature splitting speeding up model training essential for a large complex dataset like credit risk.

\textbf{4.	Hyperparameter Tuning:} The system implements hyperparameter tuning during the model training phase to enhance the efficiency of machine learning models (such as XGBoost, LightGBM, and Random Forest).Hyperparameter tuning systematically searches for the collective configuration of model parameters that yield the highest evaluation metrics (like ROC-AUC, F1-score, etc.).This is done using GridSearchCV that helps set optimal hyperparameter values for a model. Pre-define values are provided to this function in form of dictionary for each model, and GridSearchCV tries all combination for values passed. The system evaluates each combination of hyperparameters using cross-validation to ensure robust performance. The best-performing hyperparameter set is according to the selected evaluation metric (e.g. ROC-AUC). The final model is retrained using these optimal hyperparameters.

\textbf{5.	SMOTE (Class Imbalance Handling):} In classification models, often times the classes or categories are not balanced. Number of instances in class A might be more (Majority Class) than no. of instances in class B (Minority Class). This imbalance leads to biased training of the model. Model is not able to learn much about the minority class. 

This problem is fixed by generating synthetic samples of the minority class to equalise the class distribution known as Synthetic Minority Oversampling Technique (SMOTE). It automatically recognises the minority class and uses nearest neighbours in the feature space to create instances that are similar to original instances (not exact copies). The amount of oversampling is controlled to reach the same size as of majority class.

\textbf{D.	Evaluation Metrics}

Evaluation metrics give a quantitative measure to assess and understand the performance of a model. Several metrics are used according to the task performed to evaluate performance of a model. It also helps compare different models to select the best performing model or if underperforming we can tune its parameters (settings) to enhance performance.
Metrics Used:

\textbf{1.	Confusion Matrix:} It is a matrix that helps understand performance of models in classification problems by comparing predicted values with actual values for each class.

\textbf{2.	Accuracy:} Accuracy tells us how correctly the model is able to predict any given class. It gives a proportion of instances which were classified correctly out of all instances.

\textbf{3.	Precision: }Precision evaluates how correctly it is able to predict the positive cases. It checks how many positive predictions were actually positive. 

\textbf{4.	Recall:} also called Sensitivity, checks the proportion of actual positive instances identified by model out of all actual positive instances. Model can misclassify some actual positive cases and negative cases – (False Negative). It is also called True Positive Rate. 

\textbf{5.	ROC - AUC:} Receiver Operating Characteristics is a graphical representation to visualise and quantify how well a model differentiates between positive and negative cases at various thresholds. It is a trade-off between True Positive Rate/Recall (TPR) and False Positive Rate (FPR). FPR gives proportion of actual negative instances that were “incorrectly classified as positive by the model.” If curve rises towards the top left corner of the graph as represented it represents a good model.

Business Impact is also measured using approval rate, default rate , FPR and false negative rate (FNR) that evaluates how business is impacted with model’s decision of each applicant. False negative rate is crucial for reducing financial losses due to defaults. FPR is beneficial in terms to evaluate customer satisfaction and business opportunity.

\textbf{E.	Risk Assessment}

The trained model is integrated with a credit risk decision engine for risk categorisation. As per user requirements, thresholds are defined to categorise applicants into the  categories of Low, Moderate and High Risk bands. The system generates a probability score to indicate likelihood of default. This is derived from confidence score given by the models during training phase. This score is based over the feature relationships learnt during training.
These categorises ae mapped to loan decisions of whether to approve loan , review applicant or to reject applicant respectively. These rules can be adjusted as per the lender rules \& requirement.
The risk assessment engine is capable of calculating interest rates based on  the risk categories using financial calculations as per financial business rule and formulate custom loan conditions via mapping with risk categories. These loan term are rule-based and based upon the loan amount requested.

 \begin{figure}[H]
     \centering
     \includegraphics[width=0.75\linewidth]{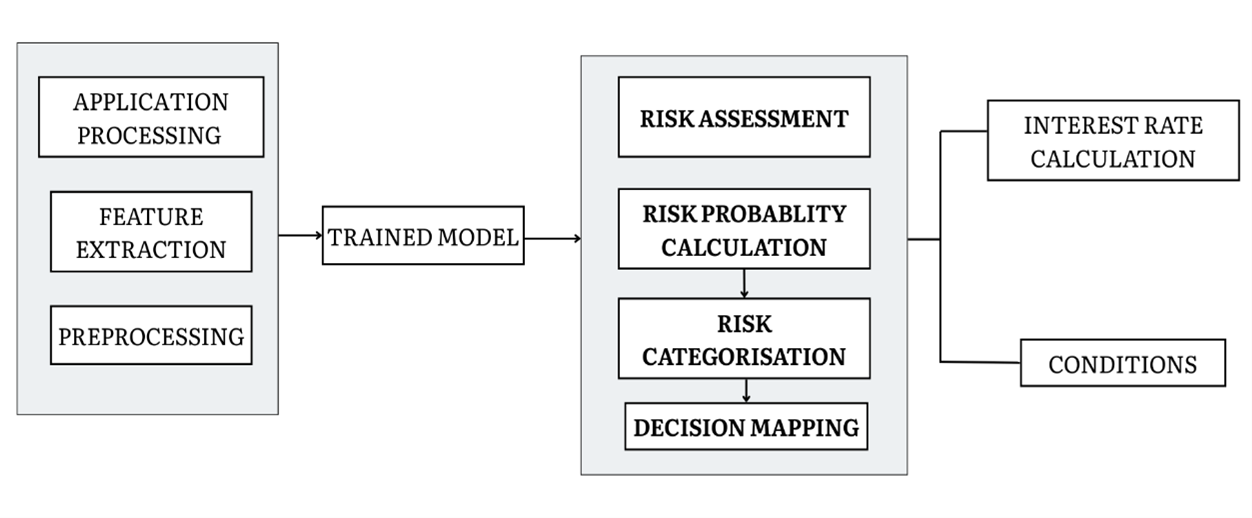}
     \caption{Risk Assessment Flow}
     \label{fig:enter-label-1}
 \end{figure}

\textbf{F.	Explainability using SHAP and LIME}

Explainable AI (XAI) refers to techniques that allows human users to comprehend and trust the results predicted  by algorithms in machine learning. XAI techniques provide insights into how AI models arrive at their predictions or actions. Models like XGBoost, LightGBM and Random Forest are “black-box” models that are highly accurate but do not provide much information about the complex patterns the model captures. This explainability is vital for building trust, ensuring accountability, facilitating debugging, and enabling effective human-AI collaboration, particularly in sensitive applications like finance and healthcare.\cite{dnv_ai}

\textbf{1.	SHAP :} SHAP is a technique / algorithm in XAI that helps interpret the prediction made by the ML model. It tells how much each feature has contributed in predicting an outcome. It is model-agnostic i.e. it works well with any ML algorithm. The SHAP outcomes are visualised through summary (beeswarm) plot and features bar plot. Values are assigned  to each feature known as the SHAP value that quantify the contribution of each feature in model prediction.

        \textbf{a.	Summary Plot: }“Each instance the given explanation is represented by a single dot on each feature row. The x position of the dot is determined by the SHAP value of that feature, and dots pile up along each feature row to show density.”\cite{shap_beeswarm}

        \textbf{b.	Bar Plot:} A global feature importance is visualised through bar plot. A matrix containing the average absolute SHAP values is input into the bar plot function to generate a visual of overall feature importance.

\textbf{2.	LIME}: LIME is another model-agnostic model to explain result of an individual data point by the model. LIME is used to explain outcome of a specific instance/data, offering a fast explanation of how a specific prediction was reached.  It provides a quick interpretability of a single prediction into how that particular prediction was made.
        a.	 LIME feature importance plot helps visualise the features supporting the model’s prediction. Their contribution is evaluated using weights assigned to each feature. Positive features are ideally represented in green indicating how model’s prediction was pushed to positive class and negative features are depicted in red indicating how model’s prediction was pushed towards negative class. Each applicants decision is supported using this LIME Feature Importance providing personalised explanation for model’s decision.

\textbf{G.	Report Generation}

Several  reports are created at the end that summarises the results and outputs of risk assessment and explainable AI. Reports are created using HTML and JSON. Different reports created are:

\textbf{•	Applicant Report}: A descriptive report about risk assessment of a loan applicant with model’s decision along with explanations about the decision with business metric. There are demo reports that are user-friendly and can be easily understood by stakeholders/applicant for review. Reports are generated for each case and stored onto the system.

\textbf{•	Business Impact Report:} A report giving an analysis on how business (lender) might be affected following the model’s decision and its impact on business in future.

\textbf{•	XAI Report:} Gives an overview and insight into model’s decision making with visualisations such as LIME Feature Importance, SHAP Analysis and” Summary plots.

\section{Results}

\textbf{A.	Feature Importance SHAP \& LIME}

SHAP Summary provides an insight about model's decision. This allows users to pick the model that suits the criteria as per users model. SHAP Summary of all three models highlights that external credit risk reports serve as an impactful parameter in assessing credit risk of an applicant. External sources are the primary indicator in evaluating credit risk.

Secondary indicator in risk assessment is the credit to goods ratio suggesting that when loan amount is larger than the asset evaluation the applicant is marked to be risky. Random Forest indicate young borrowers are potentially risky thus denying loans to major of young applicants. LightGBM emphasises on the count of family members and type of housing of applicant in determining their credit risk. XGBoost forms a decision by assessing total housing space and length of employment of the applicant as secondary parameters in classifying risk.

XGBoost and LightGBM suggest better performance stemming from the identification of sophisticated feature interactions. Common Core Dependence: All models are significantly reliant on external credit score characteristics for risk forecasting. Interpretability Value: SHAP plots render immediate insight into the decision-making of every model.

\begin{table}[htbp]
\centering
\caption{Top Feature Ranking for XGBoost, LightGBM, and Random Forest}
\resizebox{\textwidth}{!}{%
\begin{tabular}{|c|l|l|l|}
\hline
\textbf{Rank} & \textbf{XGBoost (XGB)} & \textbf{LightGBM (LGB)} & \textbf{Random Forest (RF)} \\
\hline
1 & EXT\_SOURCE\_3 & EXT\_SOURCE\_2 & EXT\_SOURCE\_2 \\
2 & EXT\_SOURCE\_2 & EXT\_SOURCE\_3 & EXT\_SOURCE\_3 \\
3 & EXT\_SOURCE\_1 & CNT\_FAM\_MEMBERS & EXT\_SOURCE\_1 \\
4 & CREDIT\_TO\_GOODS\_RATIO & YEARS\_BEGINEXPLUATATION\_MEDI & CREDIT\_TO\_GOODS\_RATIO \\
5 & TOTALAREA\_MODE & CREDIT\_TO\_GOODS\_RATIO & AGE\_YEARS \\
\hline
\end{tabular}%
}
\label{tab:feature_ranking}
\end{table}

\begin{figure}[H]
    \centering
    \includegraphics[width=0.75\linewidth]{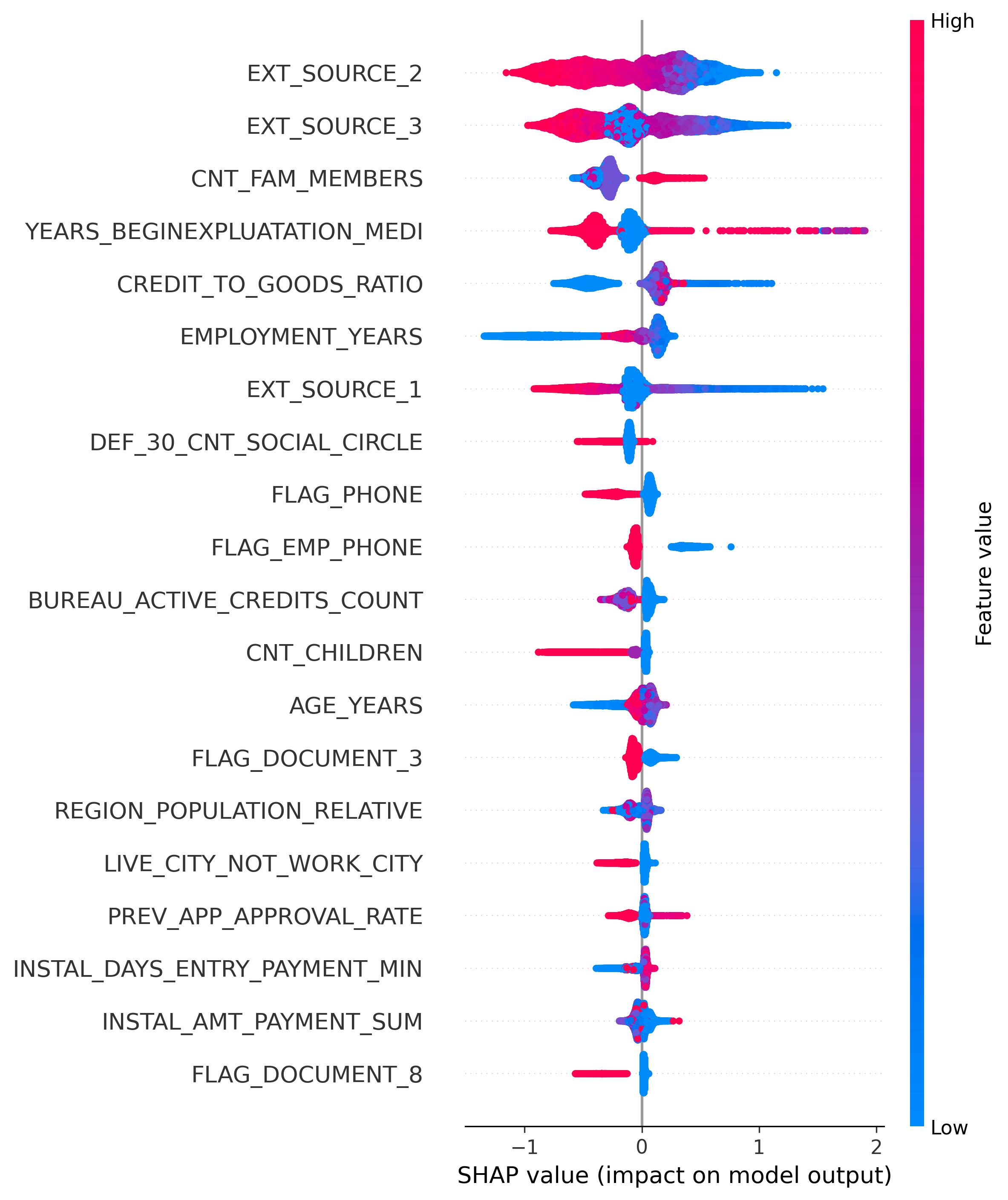}
    \caption{SHAP Summary of LighGBM (Top Performer)}
    \label{fig:enter-label-2}
\end{figure}

LIME Feature Importance gives insights into particular instances (specific applications) providing users transparency about the model's decision for each applicant. In the risk report generated for the applicants, these insights help understand model's interpretability of the risk decision. This provides decision-makers, as well as applicants, ability to review their position and reasoning.

While SHAP summaries offers users the global understanding of most impactful parameters, however it might not be the case with every applicant. LIME feature importance personalises the report,significantly reducing time for decision-makers to make a decision on the applicant's loan. Below is a sample from one of high risk customer, identified by LightGBM. The green bars give a positive impact to the application whereas red bars show factors affecting the application negatively.

\begin{figure}[H]
    \centering
    \includegraphics[width=0.75\linewidth]{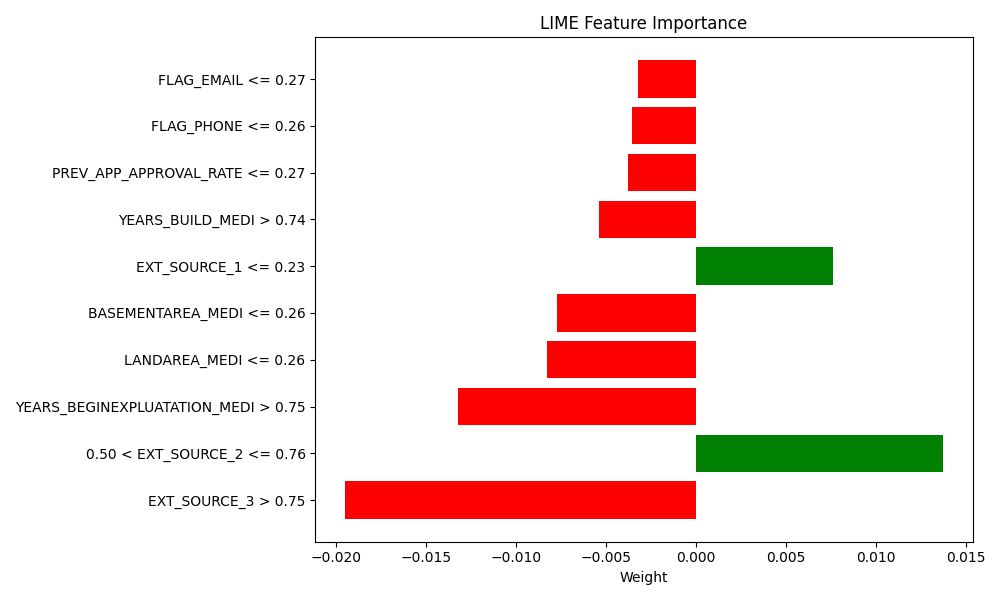}
    \caption{LIME Feature Importance of High-Risk Applicant by LightGBM}
    \label{fig:enter-label-3}
\end{figure}

\textbf{B.	Evaluation Metrics}

    \textbf{1.	Accuracy:} LightGBM has a good classification accuracy of 90.07\% for loan requests. However, in the case of imbalanced datasets, high accuracy can be misleading if the model mostly predicts the majority class (non-default).

    \textbf{2.	Precision: }This metric shows the proportion of high-risk candidates correctly predicted who are indeed high-risk. LightGBM has the best precision (0.2757), i.e., if it predicts a candidate as high-risk, it is more likely to occur as compared to the other models. Precision values are, however, mostly low for all the models, i.e., most of those predicted as high-risk are actually low-risk (Type I error).

    \textbf{3.	Recall:} The measure expresses the fraction classified as genuine high-risk candidates that are marked by the model. The random forest yields the highest recall value (0.3040) and indicates capturing a bigger ratio of genuinely high-risk entities at the expense of lower precision. LightGBM and XGBoost rank in lower recall with the indication they miss out on a larger portion of real high-risk applicants (Type II).

    \textbf{4. ROC-AUC}ROC AUC stands for “Area Under Receiver Operating Characteristic curve” and is a more sophisticated evaluation of the ability of the model to separate positive (default) and negative (non-default) classifications at different thresholds(points). The large ROC AUC of LightGBM is a measure of an improved capacity to rank candidates by their risk profiles.

        \begin{table}[h]
\centering
\caption{Evaluation Metrics}
\label{tab:evaluation}
\begin{tabular}{|l|c|c|c|c|}
\hline
\textbf{Model} & \textbf{Accuracy} & \textbf{Precision} & \textbf{Recall} & \textbf{ROC AUC} \\
\hline
LightGBM      & 90.07\%           & 0.2757             & 0.1434          & 0.7203           \\
\hline
XGBoost       & 88.74\%           & 0.2199             & 0.1566          & 0.7033           \\
\hline
Random Forest & 82.03\%           & 0.1652             & 0.3040          & 0.6808           \\
\hline
\end{tabular}
\end{table}

    \begin{figure}[H]
        \centering
        \includegraphics[width=0.75\linewidth]{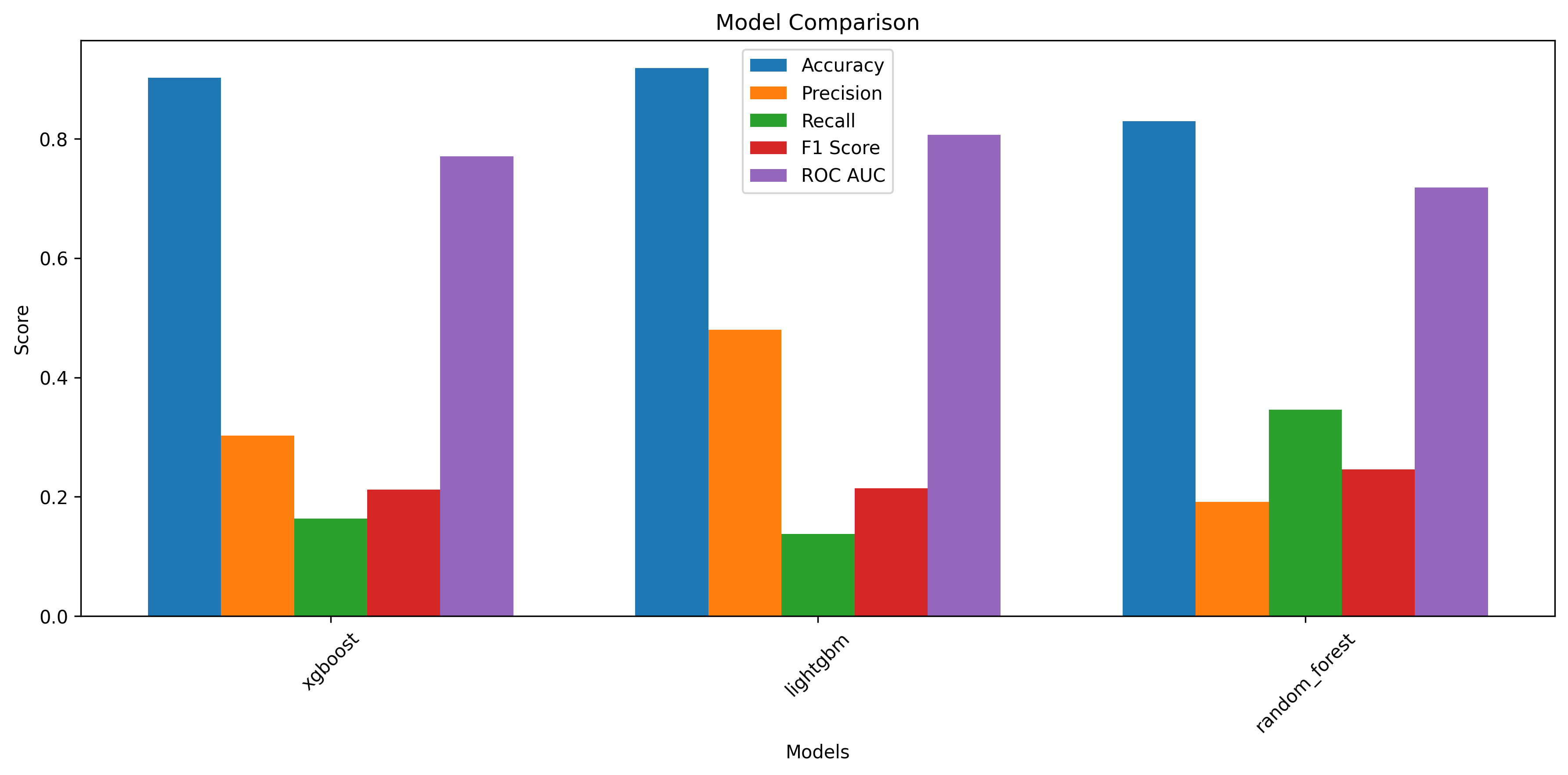}
        \caption{Evaluation metrics by model}
        \label{fig:enter-label-4}
    \end{figure}

    \begin{figure}[H]
        \centering
        \includegraphics[width=0.75\linewidth]{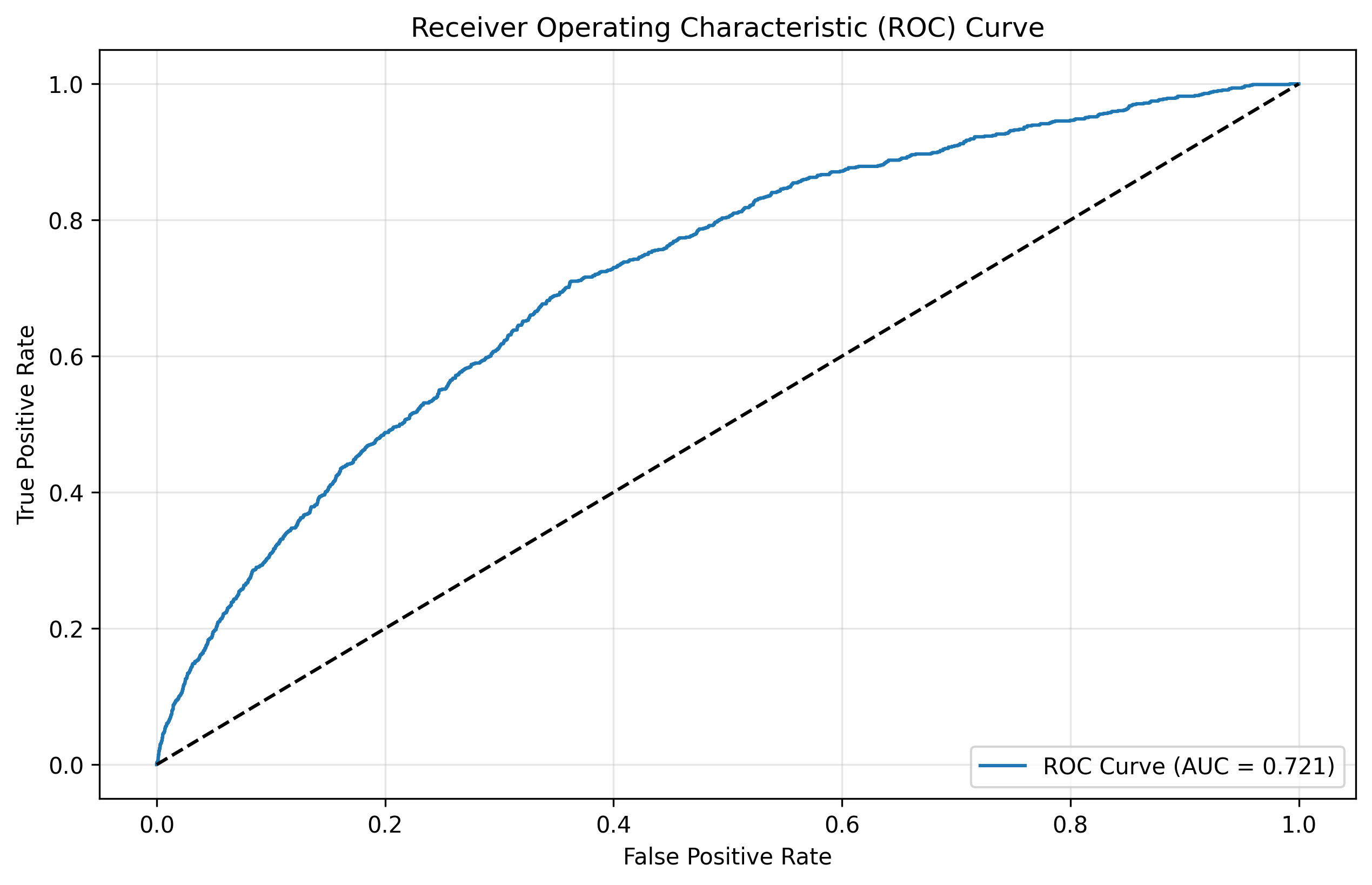}
        \caption{ROC AUC for LightGBM}
        \label{fig:enter-label-5}
    \end{figure}

    \begin{figure}[H]
        \centering
        \includegraphics[width=0.75\linewidth]{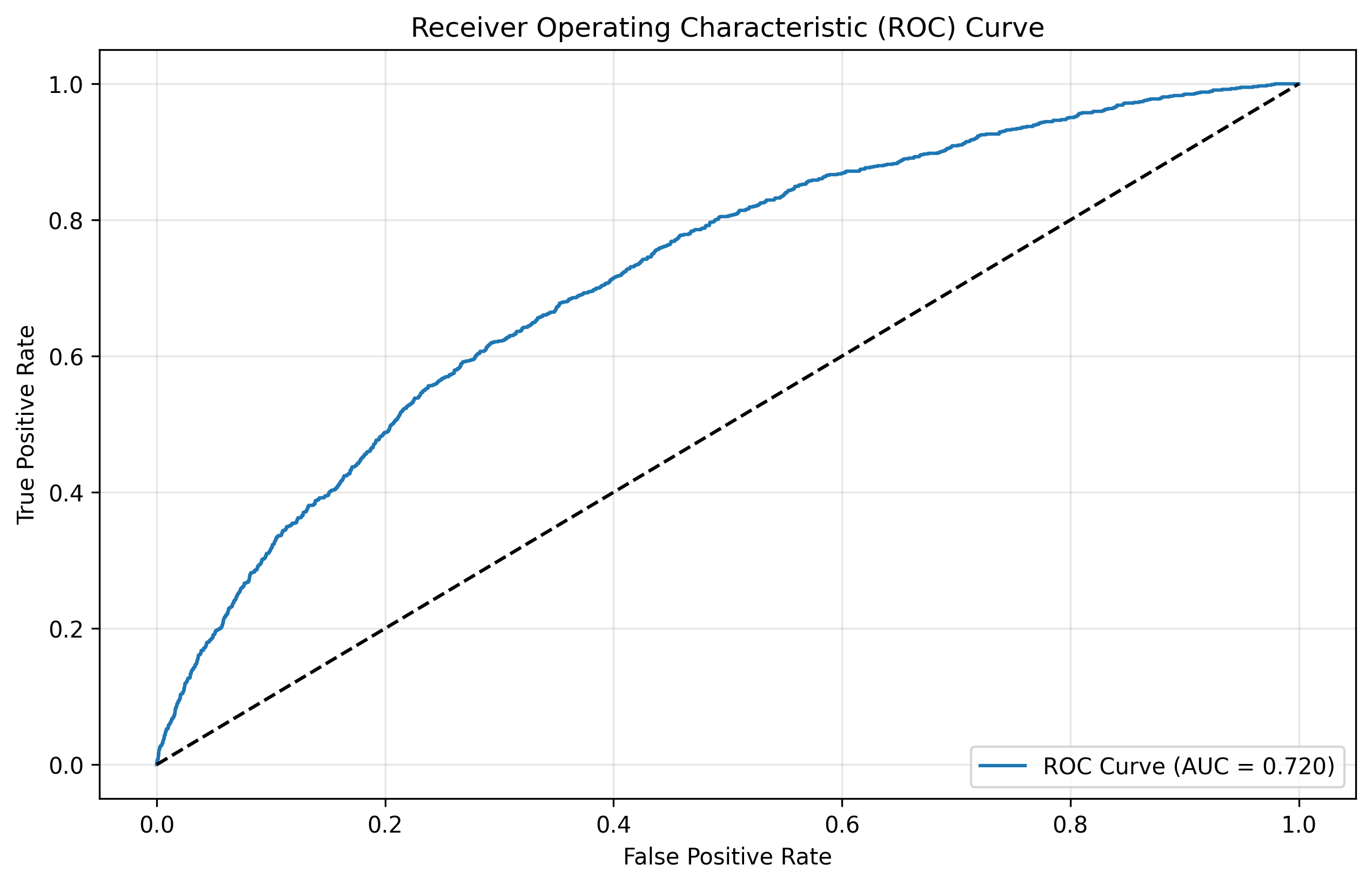}
        \caption{ROC AUC for XGBoost}
        \label{fig:enter-label-6}
    \end{figure}

    \begin{figure}[H]
        \centering
        \includegraphics[width=0.75\linewidth]{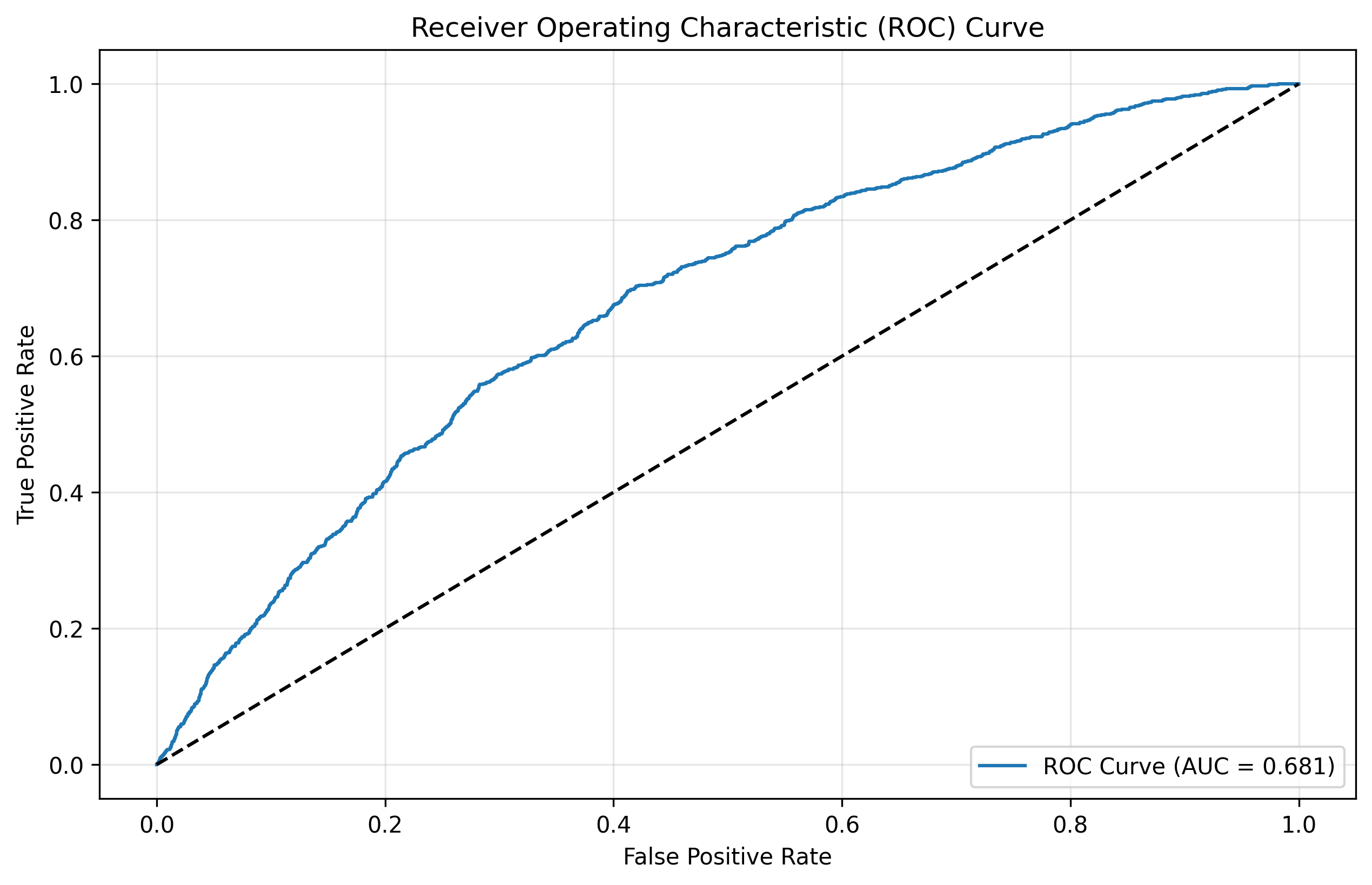}
        \caption{ROC AUC for RF}
        \label{fig:enter-label-7}
    \end{figure}

\textbf{C.	Business Metrics}
Overall, LightGBM is most likely to sanction loan requests and is associated with lowest proportion of defaults among all approved loans. LightGBM is also effective in minimizing the false positives as well as false negatives. This is beneficial for customer satisfaction and business opportunity while reducing financial losses due to defaults. XGBoost performs fairly as the approval rate is comparable with LightGBM but the approved loans have a higher chances of defaulting. In terms of customer satisfaction and business opportunity this model gives almost satisfactory results but is prone to encountering financial loses. The worst in terms of business opportunity is random forest as it is prone to encounter heavy financial loss even though it is cautious in approving loans.

 \begin{figure}[H]
     \centering
     \includegraphics[width=0.5\linewidth]{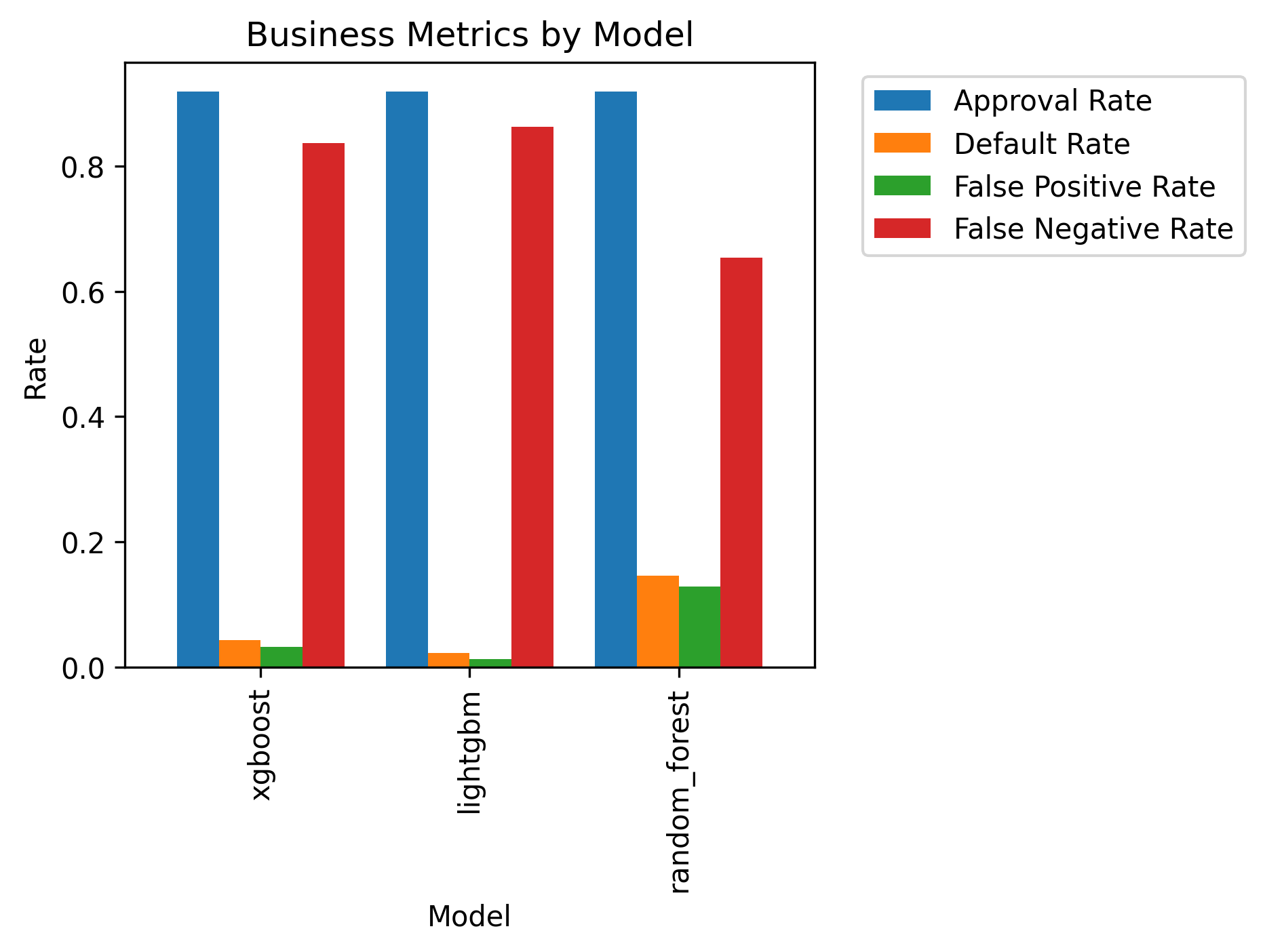}
     \caption{Business metrics by model}
     \label{fig:enter-label-8}
 \end{figure}

\textbf{D.	Top Classifier}
 LightGBM shows top performance in terms of accuracy (90.07 \%), business impact, approval rates and ROC-AUC. XGBoost performs near to LightGBM in terms of accuracy (88.74\%) and approval rate but gives fair performance when considering business impact of the results generated by XGBoost. LightGBM is optimized for speed and accuracy, hence is very reliable for the provided dataset. Using Random Forest (82.03\%) comes with lots pf negative factors with the lowest accuracy among the three models trained. On all parameters. Random Forest underperforms hence is unreliable when used for credit risk assessment.

\section{Conclusion}
This multi-model system uses three ML algorithms XGBoost, LightGBM and Random Forest. LightGBM achieves the highest accuracy (90.07\%) suggesting correct classification of majority of applicants with highest approval rate of 95\%. Other models XGBoost (88.74\% accuracy) and Random Forest (82.03\% accuracy) also perform nearly well with LightGBM. Random Forest has best recall meaning it is most effective at catching risky applicants. This represents models' ability to distinguish well between low-risk and high-risk applicants, which translates to more precise and trustworthy automated decision-making. The system also optimizes the credit approval process by converting model predictions into actionable risk buckets and unambiguous loan decisions (Approve, Review, Reject), minimizing manual intervention and increasing efficiency. The business impact potential is significant, as the system offers a data-driven framework for lending decisions, ultimately seeking to lower default rates and improve overall profitability by precisely quantifying risk and potential financial returns.

The most remarkable novelty of the system is its simple, automated report generation. For every credit application, the system generates sophisticated and visually appealing reports that make complex model decisions understandable to a broad spectrum of stakeholders, including non-technical business users and even applicants. The strategic implementation of SHAP and LIME for “global and local interpretability” gives precise, easy-to-understand explanations for individual risk assessments and drivers of the decisions. The flexibility of the system is also enhanced by its capability to generate output in various formats, including HTML for easy consumption, PNG for easy sharing and embedding, and JSON for programmatic consumption and further integration.
In summary, this project provides a futuristic, explainable credit risk assessment system that not only streamlines and optimizes the loan approval process but also demystifies the underlying decision-making process through its innovative and user-friendly reporting features. The system's capability to make accurate predictions which are easy to understand, human-readable explanations makes it an effective tool for driving efficiency, minimizing risk, ensuring compliance, and building trust in the lending process.

\section*{Acknowledgment}
The author wishes to extend her sincere gratitude towards Mr. Harsh Pathak for enlightening and guiding her with his invaluable input and insightful feedback throughout this study.

\bibliographystyle{IEEEtran}
\bibliography{references} 
\end{document}